\documentclass{sn-jnl}% Math and Physical Sciences Reference Style
\usepackage{algorithm}
\usepackage{algpseudocode} % This package provides the pseudocode environment
\usepackage{multirow}%
\usepackage{amsmath,amssymb}%
\PassOptionsToPackage{dvipsnames}{xcolor}
\usepackage{lscape}

\jyear{2023}%

\begin{document}
\title{AICAttack: Adversarial Image Captioning Attack with Attention-Based Optimization}
% \title{AICAttack}
%%=============================================================%%
%% Prefix   -> \pfx{Dr}
%% GivenName    -> \fnm{Joergen W.}
%% Particle -> \spfx{van der} -> surname prefix
%% FamilyName   -> \sur{Ploeg}
%% Suffix   -> \sfx{IV}
%% NatureName   -> \tanm{Poet Laureate} -> Title after name
%% Degrees  -> \dgr{MSc, PhD}
\author[1]{\fnm{Jiyao} \sur{Li}}\email{jiyao.li-1@student.uts.edu.au}

\author[1]{\fnm{Mingze} \sur{Ni}}\email{mingze.ni@uts.edu.au}
\author[1]{\fnm{Yifei} \sur{Dong}}\email{yifei.dong@uts.edu.au}

\author[2]{\fnm{Tianqing} \sur{Zhu}}\email{tqzhu@cityu.edu.mo}
\author[3]{\fnm{Yongshun} \sur{Gong}}\email{ysgong@sdu.edu.cn}
\author*[1]{\fnm{Wei} \sur{Liu}}\email{wei.liu@uts.edu.au}
\affil[1]{\orgdiv{School of Computer Science}, \orgname{University of Technology Sydney}, \orgaddress{\street{15 Broadway}, \city{Sydney}, \postcode{2007}, \state{NSW}, \country{Australia}}}
\affil[2]{\orgdiv{Faculty of Data Science}, \orgname{City University of Macau}, \orgaddress{\street{81 Av. Xian Xing Hai}, \city{Macau}, \postcode{999078}, \state{Macau}, \country{China}}}
\affil[3]{\orgdiv{School of Software}, \orgname{Shandong University}, \orgaddress{\street{27 Shanda Nanlu}, \city{Jinan}, \postcode{250100}, \state{Shandong}, \country{China}}}

\abstract{Recent advances in deep learning research have shown remarkable achievements across many tasks in computer vision (CV) and natural language processing (NLP). At the intersection of CV and NLP is the problem of image captioning, where the related models' robustness against adversarial attacks has not been well studied. This paper presents a novel adversarial attack strategy, AICAttack (Attention-based Image Captioning Attack), designed to attack image captioning models through subtle perturbations on images. Operating within a black-box attack scenario, our algorithm requires no access to the target model's architecture, parameters, or gradient information. We introduce an attention-based candidate selection mechanism that identifies the optimal pixels to attack, followed by a customized differential evolution method to optimize the perturbations of pixels' RGB values. We demonstrate AICAttack's effectiveness through extensive experiments on benchmark datasets against multiple victim models. The experimental results demonstrate that our method outperforms current leading-edge techniques by achieving consistently higher attack success rates.}

\keywords{Adversarial Learning, Adversarial Attacks, Computer Vision, Neural Language Processing, Image Captioning}



\maketitle

\section{Introduction}

In recent years, deep learning models, particularly Convolutional Neural Networks (CNNs), have showcased remarkable achievements across diverse computer vision tasks, notably image classification. These models have attained human-level or surpassed human performance \cite{cite8}, thus opening avenues for their practical integration into real-world applications. Nevertheless, this swift advancement has brought to light a critical vulnerability inherent in these models - their susceptibility to adversarial attacks.

In computer vision tasks, adversarial attacks aim to introduce meticulously crafted perturbations to input images, thereby causing models to yield erroneous or misleading predictions \cite{cite3}. These perturbations can profoundly influence model outputs despite being imperceptible to human observers. Adversarial image attacks predominantly target tasks rooted in CNNs, with image classification as common examples \cite{cite1,cite2,cite3}.

Examining attacks in image classification tasks from an input perspective, the conventional approach involves injecting perturbations into the original image to prompt the model to generate an incorrect classification label. Computations involving gradients often play a crucial part in directing attacks aimed at image classification problems. In the context of white-box attacks, access to the model's gradient is feasible, allowing researchers to derive perturbations by minimising the redefined objective function \cite{cite9}.
% \begin{figure}[t]
%     \centering
%     \includegraphics[width=0.47\textwidth]{shallow_example.png}
%     \caption{An example of our captioning attack performed on Show Attend and Tell \cite{cite23}, with our AICAttack approach.}
%     \label{fig:an_iamge_demo}
% \end{figure}
Image captioning represents a closely related domain, encompassing the generation of coherent and intelligible captions for images through meticulous analysis. The predominant attacking methodology to captioning entails the deployment of CNNs to extract image features and RNNs (Recurrent Neural Networks) to formulate descriptive captions \cite{cite4, cite5, cite6, cite7}, which is commonly denoted as an Encoder-Decoder architecture.

Crafting adversarial attacks against image captioning models poses unique challenges that surpass those in image classification attacks. These difficulties primarily arise from the complexities of leveraging gradients within the Encoder-Decoder framework \cite{cite33}. The field's main hurdles can be distilled into two key issues: the impracticality of using internal model information for attacks, and the intricacy of accurately evaluating attack effectiveness on generated text \cite{cite25,cite38}.

Most existing studies on adversarial attacks targeting image captioning systems have concentrated on white-box scenarios \cite{cite30}, assuming complete attacker knowledge of the model's architecture and parameters. However, this assumption often proves unrealistic in practical settings, where attackers rarely have comprehensive access to the target model's inner workings. 

\textbf{Motivation for this research:} As image captioning models are now widely utilized in real-world applications such as accessibility \cite{cite41, cite42}, journalism \cite{cite39, cite40}, and autonomous systems \cite{cite43, cite44}, these models remain susceptible to adversarial attacks, leading to incorrect or misleading image captions in critical contexts. This emphasizes the need to thoroughly examine how adversarial examples can compromise image captioning models and explore strategies to defend against these threats—an area that has not been fully investigated. To address this disconnect between theoretical assumptions and real-world applications, we introduce a novel approach that better reflects authentic adversarial conditions.

Our proposed methodology, AICAttack (Attention-
based Image Captioning Attack), integrates an attention mechanism to precisely identify and target the most susceptible pixels in an image for adversarial manipulation. We then utilize a differential evolution algorithm to optimize the attack's effectiveness, ensuring that the generated adversarial samples are impactful and plausible. This strategy not only tackles the practical constraints of previous methods but also enhances the viability and applicability of adversarial attacks in image captioning.

Our work makes the following key contributions:
\begin{itemize}
    \item We present AICAttack, a novel adversarial attack method employing an attention mechanism to accurately locate the pixels most critical to caption generation. This strategy enables us to target our efforts on areas with the greatest potential to influence captions, thus enhancing attack efficiency without relying on gradient computations.
    \item Our approach is further refined by a differential evolution algorithm. This algorithm is customized to precisely adjust adversarial modifications on the identified key pixels, ensuring the alterations are both imperceptible and highly effective.
    \item We perform extensive testing across diverse real-world datasets, targeting various image captioning models as potential victims. These comprehensive experiments demonstrate the efficacy of our approach in creating adversarial examples that effectively undermine the reliability of image captioning systems.
\end{itemize}
The structure of this paper is as follows. Section 1 introduces the background of this study and briefly states our research contribution. Section 2 on related work discusses existing research and highlights how our study contributes to the existing knowledge. Section 3 introduces our proposed attack method, AICAttack, in detail. Section 4 provides experiments and analyses that validate our proposed method. Section 5 concludes this research and provides directions for our future work.

\section{Related Work}  
\label{sec: related work}
With the advent of the fast gradient sign method (FGSM) \cite{cite9}, deep neural network models (DNNs) have exhibited vulnerabilities to adversarial examples. A multitude of gradient-based attack techniques have emerged, such as the Basic Iterative Method (BIM) \cite{cite10} and Projected Gradient Descent (PGD) \cite{cite11}, which enhanced FGSM by iteratively updating perturbations, resulting in more potent attacks. Notably, these approaches are regarded as white-box attacks, requiring access to the model's configuration for execution, which is impractical in real-world scenarios. Moreover, white-box attacks only target specific models and have relatively weak transferability \cite{yin2023transfer}.

To transcend this constraint, black-box attacks based on generative adversarial networks (GANs) have been developed \cite{cite12,cite13}. These attacks deploy GANs to generate perturbations that can mislead the model. However, training GANs is often fraught with instability. Besides, tasks that target an image, such as the image captioning task, cannot meet the requirement of modifying pixels. Another genre of attacks, optimisation-based attacks \cite{cite14,cite16}, offers a potential solution. These attacks cast the creation of adversarial examples as an optimisation problem, wherein the objective is to identify the minimal perturbation that induces misclassification \cite{cite17}. Similar attack strategies have also been studied for attacking road sign recognition \cite{yang2020targeted} and other problems with Stackelberg games \cite{yin2018sparse,chivukula2017adversarial}.

For generating image captions, state-of-the-art models typically employ an encoder-decoder architecture \cite{cite18}. This architectural framework augments the model's capacity to extract latent features by integrating elements like attention mechanisms. These components facilitate precise contextual mapping between the input image and the generated caption, substantially enhancing overall performance. Several alternative models have predominantly leveraged semantic information or compositional architectures for caption generation \cite{cite19, cite20}.

The limitation for attacking the image captioning model arises from the unique characteristics of generative models, which yield discrete sequence outputs in captions, unlike the context of image classification \cite{cite21, cite22}. The intricate interplay among words within a sentence introduces complexities in calculating gradients for the loss function within an adversarial attack scenario. To circumvent this challenge, a solution proposed by \cite{cite24} involves treating the entire sentence as an independent output during adversarial attacks.

Recent efforts have emerged focusing on manipulating generated captions of image captioning models through adversarial attacks to incorporate specific vocabulary. Two prominent approaches are \cite{cite19} and \cite{cite33}. In both \cite{cite19} and \cite{cite33}, the problem of attacking the image captioning model is defined as an optimization task with constrain (target caption). These attacks conventionally exploit the model's internal parameters, such as gradients, to generate adversarial instances. However, these methodologies exhibit distinct limitations since they necessitate access to internal model information, rendering them susceptible to detection and unsuitable for deployment in a black-box setting.

\section{Our Proposed Attack Method}
\label{sec: Methodology}
This section elaborates on our proposed image-captioning attack method, AICAttack (Attention-based Image Captioning Attack).

\begin{figure*}[t]
    \centering
    \includegraphics[width=0.95\textwidth]{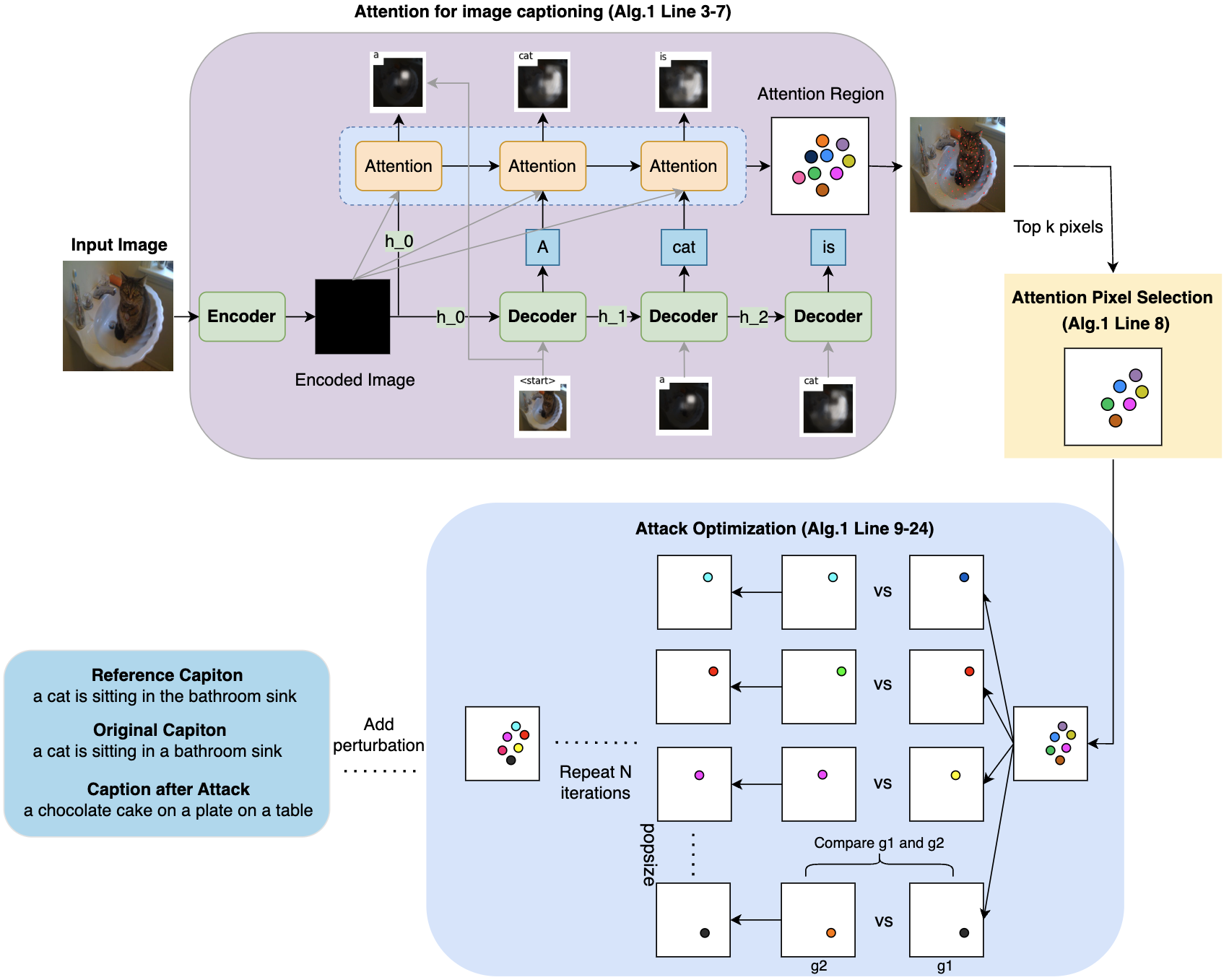}
    \caption{The Workflow of our AICAttack Algorithm for Image Captioning Attacks. The process begins by feeding the input image into the attention block, which generates attention scores. These scores are then used for attention pixel selection. During the attack optimization phase, the Differential Evolution (DE) algorithm searches for the most effective adversarial sample.}
    \label{fig:overall}
\end{figure*}

\subsection{Problem Setting}
Given a pre-trained image captioning model \( F(\cdot): \mathcal{X} \rightarrow \mathcal{Y} \), where \( \mathcal{X} \) represents the image feature space and \( \mathcal{Y} \) represents the textual space, an attacker seeks to generate an adversarial image \( \mathbf{x'} \) by manipulating an existing image \( \mathbf{x} = (x_1, \ldots, x_i, \ldots, x_P) \), where \( \mathbf{x} \in \mathcal{X} \), \( x_i \in \mathbb{R} \), and \( P \) is the number of pixels. The objective is to deceive the performance of \( F(\cdot) \) such that \( F(\mathbf{x'}) \) does not match the ground truth \( y \in \mathcal{Y} \).

To craft an adversarial example, a perturbation \( \Delta \mathbf{x} \) is added to the image \( \mathbf{x} \), resulting in the construction of the adversarial example \( \mathbf{x'} \) as follows:
\begin{equation}
\label{equation: adv calculate}
\begin{split}
    \mathbf{x'} &= \mathbf{x} + \Delta \mathbf{x},\\
    \text{where}\quad \Delta \mathbf{x} &= (\Delta x_1, \ldots, \Delta x_i, \ldots, \Delta x_P).
    % \Delta i_{c1} &= (x_1, y_1, r_1, g_1, b_1)
\end{split}
\end{equation}

In the above context, attacking the victim model \( F \) involves the process of searching for \( \Delta \mathbf{x} \). To construct such a perturbation \( \Delta \mathbf{x} \), the attacker first identifies \( m \) (where \( m \leq P \)) pixels using indices \( \mathbf{I} = \left\{ I_j \right\}^m_{j=1} \), and then optimizes the corresponding perturbation values \( \left\{ \Delta x_{I_j} \right\}_{I_j \in \mathbf{I}} \). Additionally, for the unaltered pixels, their perturbation values are set to \( 0 \) (\( \Delta x_h = 0 \) for \( h \notin \mathbf{I} \)).

In the following sections, we introduce our AICAttack algorithm in detail. An illustration figure of the AICAttack process is shown in Fig.~\ref{fig:overall}. An input image goes through an attention layer to generate attention scores. Based on that, an attack optimization step is implemented to generate the optimal adversarial example. 
% \textcolor{red}{To overcome the difficulty of
% utilizing gradients within the encoder-decoder structure, Our algorithm works in a Black-box attack scenario, assuming the adversary cannot access internal parameters and gradient information.} 
Diverging from conventional algorithms, we transform this task into a problem of discovering the optimal solution within a specified region. Therefore, we have two main tasks to accomplish: (1) select the optimal locations of pixels to be attacked. (which we detail in Sec.~\ref{Attention for Candidate Selection}), and (2) determine the optimal perturbation values for the selected pixels. (which we discuss in Sec.~\ref{Differential Evolution Optimization}).

% These tasks are addressed respectively in the following two subsections. 

%with Algorithm~\ref{algorithm: Attack Approach}. 
% To quantitatively assess the success of an attack, we employ BLEU \cite{cite27} score, denoted as $B$. In image classification tasks, the accuracy of generating incorrect labels is often used to measure successful attacks. However, in image captioning jobs, most models rely on textual evaluation metrics such as BLEU score and others. Hence, we chose BLEU to monitor our attack quality. BLEU score is calculated by comparing the n-grams of predicted text with the n-grams of the reference text. We evaluate the success of an attack by calculating the difference in BLEU scores before and after the attack. By the end of identifying the best adversary, our focus in this study primarily shifts to:
% \begin{equation}
% \label{equation: problem define}
%     maximize ~ B(F_C(I)) - B(F_C(I_{adv}))
% \end{equation}

% One of the most effective approaches to address attacks is strategically selecting the most vulnerable pixels within the image boundaries and making minimal adjustments to their pixel values. This generates model captions that differ in meaning from the original captions. Therefore, we divide the entire problem into two main parts. Firstly, identify the pixels in the image most susceptible to generating attack effects. Secondly, selecting the optimal perturbations for these identified pixels. The following section will introduce the methods used to achieve these two objectives separately:
\subsection{Attention for Candidate Selection}\label{Attention for Candidate Selection}

% Just draw a picture like a man holding a football
Attention-based networks enable models to choose only the parts of the encoded variables relevant to the task encountered. Bahdanau \cite{cite29} used it to address the challenge of handling long-range dependencies in lengthy textual sequences within natural language processing. As a type of soft attention, it uses a learned attention function to compute attention weights for each element in the input sequence. The same mechanism can be used in other models where the encoder's output has multiple points in space or time. In image captioning, specific pixels are usually assigned higher importance than others. We consider these high-importance pixels as potential candidate regions to be attacked.

\begin{figure}[t]
    \centering
        \includegraphics[width=0.6\linewidth]{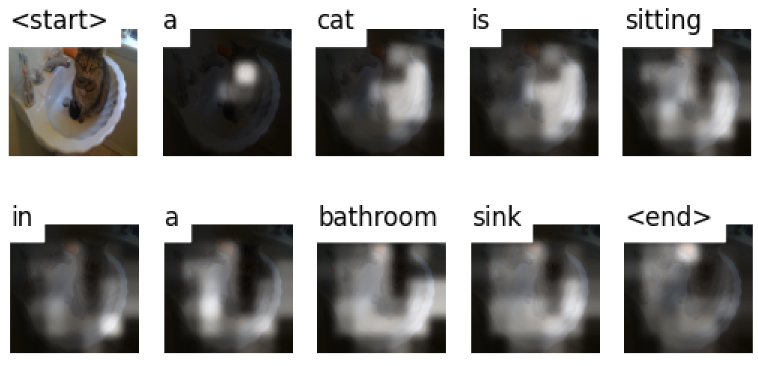}
    \caption{Attention Mechanism Illustration in a Small Cat Image Example. Highlighted regions denote attention concentration guiding the encoder-decoder network during word generation processes.}
    \label{fig:attention in step}
\end{figure}
% {0.48\textwidth}

In this research, we extracted the attention score $\alpha$ derived from an image captioning network, such as Show, attend and tell (SAT) \cite{cite23}. For each pair of input images and generated captions with a length of $l$, we derive attention mappings for each individual word in the caption. Fig.~\ref{fig:attention in step} shows a visual example of the attention mapping for each word. Given our use of soft attention, where there are $P$ pixels, and the pixel weights sum up to 1, for each word token $t$, we have:
%\begin{equation}
%\label{equation: attention add up to 1}
    $\sum_{1}^P \alpha_{p,t} = 1$, where $\alpha_{p,t}$ is the attention score of pixel $p$ for the word token $t$. 
%\end{equation}When handling attention scores, for each individual word, we possess a vector of attention scores equal to the size of the image, where each score is associated with a pixel value.

\begin{figure}[t]
    \centering
    % \hfill
\includegraphics[scale=0.43]{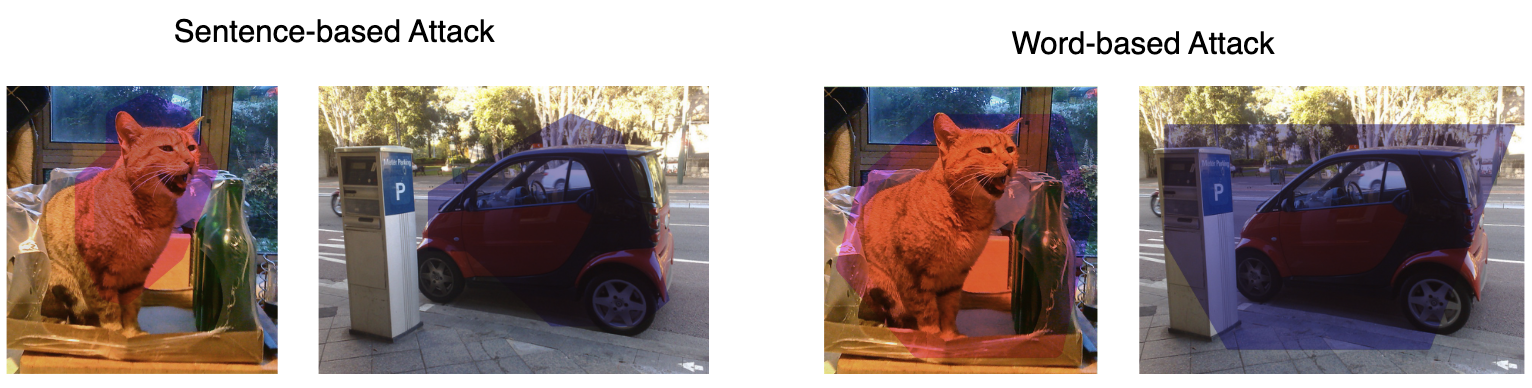}
\caption{Examples of ``Sentence-based Attack" (our proposed method) and ``Word-based Attack" approach for computing attention scores. The highlighted areas represent the attention region for pixel selections.}
% illustrating their contrasting approaches to feature detection.}
\label{fig:attention_area}
\end{figure}

Two methodologies were used to utilize the attention scores we obtained. First, ``Sentence-based Attack" involves aggregating attention scores for all pixels of all words. This results in an overall attention mapping that matches the dimensions of the original image. Subsequently, attention scores are ranked, and the pixel coordinates of the top $k$, where $k\leq P$, values are selected to form a candidate region. The second approach is referred to as ``Word-based Attack", which can be considered a baseline approach. In this case, we only focus on the top $k$ pixels ranked by attention scores in each attention mapping (each word) and then join these regions to form the candidate region. 
\textbf{Analysis of the example images: } 
The two example images in Fig.~\ref{fig:attention_area} contain main entities: a cat (left) and a car (right). 
In the ``Sentence-based Attack'', the attention region (highlighted area in the image) precisely targets the essential objects - the cat and car. In contrast, the ``Word-based Attack'' compromises its effectiveness by incorporating surrounding elements such as nearby bottles and parking meters. We argue that the ``Sentence-based'' approach achieves more efficient attention utilization by focusing on critical elements, while the ``Word-based'' method's broader attention region weakens attack performance by including non-essential regions.
The AICAttack algorithm is shown in Algorithm~\ref{algorithm: Attack Approach}, while the candidate region formulation is shown in line 6 of the algorithm.

% This optimization problem involves finding the ideal $(x, y, r, g, b)$ values for each targeted pixel. 
% There are many well-established optimization algorithms such as FGSM (Fast Gradient Sign Method) \cite{cite9} or PGD (Projected Gradient Descent) \cite{cite11}.

\subsection{Differential Evolution Optimization}\label{Differential Evolution Optimization}
After selecting the targeted pixels, our next step is to derive the best magnitudes of attacks on the pixels. We want the magnitude of the attack to be as small as possible while its impacts on the generation of captions as much as possible. To do so, we optimize the pixel perturbation values to decrease the caption's quality measured by BLEU. To this end, we apply Differential Evolution (DE) \cite{cite17}, a robust evolutionary algorithm, to solve this optimisation problem. 
DE maintains a population of candidate solutions, often called individuals. The key idea behind it is the differential mutation operator, which involves creating new candidate solutions by perturbing the difference between two other solutions from the population. New solutions are generated through mutation operation, and their fitness is evaluated. If a new solution outperforms its parent, it replaces the parent in the population.
The advantage of Differential Evolution to this problem lies:
\begin{itemize}
    \item Global Optimization Capability: DE operates on a population of solutions, facilitating a comprehensive search for optimal perturbations. It is more effective at avoiding suboptimal solutions and less prone to get stuck in local minima compared to gradient-based algorithms like FGSM (Fast Gradient Sign Method) \cite{cite9} or PGD (Projected Gradient Descent) \cite{cite11}, which often struggle with such issues.

    \item Flexibility: The problem of optimizing $x, y, r, g, b$ involves adjusting the pixel locations and RGB color values, which are discrete variables. Differential Evolution (DE) does not require the optimization problem to be differentiable, unlike traditional methods like gradient descent which is suitable for discrete variables.
    \item Robustness to Model Architecture: DE can be applied regardless of the model type or architecture, while FGSM and PGD are typically tied to specific architectures due to their reliance on gradients. This is crucial in generating image adversary since Calculating gradient requires much more information about the target system, which can be hardly realistic in many cases.
\end{itemize}
% DE's simplicity and versatility have made it popular for optimising many problems, including those with continuous, discrete, or mixed-variable spaces. Its effectiveness lies in its ability to navigate complex search spaces and uncover optimal or near-optimal solutions through iterative adaptation and evolution. 
% Differential Evolution (DE) offers the following two advantages:
% \begin{itemize}
%     \item Simplicity and Versatility: DE's straightforward implementation and minimal parameter tuning make applying it to various optimisation tasks easy. Differential Evolution (DE) does not demand a differentiable optimisation problem, a prerequisite imposed by classical optimisation techniques like gradient descent methods. \cite{cite17} Its versatility allows it to handle issues with continuous, discrete, or mixed-variable domains, making it suitable for a wide range of applications.
%     \item Global and Local Search: DE's differential mutation mechanism enables it to perform both global and local search efficiently. By introducing diversity through mutation and selecting promising solutions based on fitness, DE can effectively navigate different regions of the search space, helping to locate optimal or near-optimal solutions.
% \end{itemize}

\begin{algorithm}[t]
\caption{AICAttack: Our Proposed Adversarial Image Captioning Attack}
\label{algorithm: Attack Approach}

\begin{algorithmic}[1]
    \State \textbf{Input:} Captioning model $F(\cdot)$, image $\mathbf{x}$, attention network $A$, number of pixels $P$, attention region size $k$, population size $popsize$, iteration time $T$, attacking strength $s$, BLEU score calculation function $B$
    \State \textbf{Output:} Optimal adversarial sample $\mathbf{x'}$
    \State \text{/* Attention for Candidate Selection */}
    \State $\alpha \gets A(I)$, $ \alpha^* \gets [\ ]$
    \For{each $x$ in $\alpha$}
        \State $\alpha^* \gets \alpha^* + x$
    \EndFor
    \State Pick top-$k$ pixels from $\alpha^*$
    \State \text{/* Differential Evolution Optimization */}
    \For{$i = 1$ to $popsize$}
        \State Construct $\mathbf{x}^{i}$ where pixel locations and changes are determined by attention weights and attack strength, respectively
    \EndFor
    
    \State $\mathbf{x}^0 \gets \mathbf{x}$, $\mathbf{x'} \gets \mathbf{x}^0$
    
    \For{$g = 1$ to $T$}
        \For{$j = 1$ to $popsize$}
            \State Build $\mathbf{x}_j^g$ from the previous generation $\mathbf{x}^{g-1}$ using mutation
            \If{$B(F(\mathbf{x}^g_j)) < B(F(\mathbf{x'}))$}
                \If{$B(F(\mathbf{x}^g_j)) < B(F(\mathbf{x}^{g - 1}_j))$}
                    \State $\mathbf{x'} \gets \mathbf{x}^{g}_j$
                \EndIf
            \EndIf
        \EndFor
    \EndFor
    
    \State \Return The best attack example $\mathbf{x'}$
\end{algorithmic}
\end{algorithm}

In this research, we customize DE to obtain an optimal solution by finding the best pixel coordinates and RGB values to attack a given input image. Each candidate solution's perturbation encompasses the coordinates/locations of pixels and the changes in the pixels' RGB values. In our configuration, the initial count of candidate solutions (population) is set to $popsize$ (which is a parameter that can be changed to adapt to different applications and scenarios). Accordingly, by the DE algorithm, 
% we Construct initial generation $X_i$ with $x, y$ values confirmed by attention regions and $r, g, b$ values generate by $s$, then 
every new iteration generates $popsize$ new candidate solutions (children candidates) according to Equation~\ref{equation: new population}:
\begin{equation}
    \label{equation: new population}
    \begin{split}
    &\mathbf{x}_j^g \leftarrow \mathbf{x}^{g-1}_{r_1} + \lambda\cdot(\mathbf{x}^{g-1}_{r_2}-\mathbf{x}^{g-1}_{r_3})\\
    &where\quad r_1\neq r_2\neq r_3
    \end{split}
\end{equation}
\noindent where $\mathbf{x}_j^g$ is the candidate solution, $g$ and $j$ represent the indices of generation and the mutant in population, respectively. $\lambda$ is a parameter for candidates weight balancing and $r_1,r_2,r_3$ are random positive integers.

% The complete procedure of our method is presented in Algorithm~\ref{algorithm: Attack Approach}. 
After the attention-based candidate selection (lines 3 to 7 of Algorithm~\ref{algorithm: Attack Approach}), the algorithm initialises a population of candidate solutions (lines 9 to 13 of Algorithm~\ref{algorithm: Attack Approach}), where each solution represents a perturbed image. The DE algorithm then iteratively updates these solutions by performing differential mutation and crossover operations (lines 14 to 23 of Algorithm~\ref{algorithm: Attack Approach}). For each generation, the algorithm evaluates the fitness of candidate solutions using the BLEU score calculated with its predicted caption and compares it to the previous generation. If a candidate solution yields a lower captioning BLEU score (indicating success in fooling the victim model), it is selected as the new adversarial example.
% Key Parameters in the algorithm:
% \begin{itemize}
%     \item Population Size ($popsize$): Determines the number of candidate solutions in the population, influencing the diversity and exploration of the optimisation process.
%     \item Iteration Time ($T$): Specifies the number of iterations or generations for the DE algorithm to evolve the candidate solutions.
%     \item Scale Parameter ($mut$): Controls the scale of perturbation adjustments in the DE algorithm, influencing the search space exploration. We set it to be 0.5.
%     \item Attacking Strength ($s$): Defines the extent of perturbation applied to the RGB colour values, influencing the intensity of the attack.
% \end{itemize}
% To examine the attacking strength, we set a parameter of strength denoted as $s$, which we will study and evaluate in the next section.

\section{Experiment and Analysis}\label{sec:experiment}

In this section, we comprehensively evaluate the performance of our method against the current state of the art. Besides the main results of attack performance and imperceptibility (Sec.~\ref{experiment analysis}), we also conduct experiments on ablation studies (Sec.~\ref{Ablation Studies}), transferability (Sec.~\ref{Transferability of Attacks}), adversarial retraining (Sec.~\ref{Adversarial Retraining}). We provide code for reproductivity of our experiments\footnote{We provide our code in an anonymous setting for review: \burl{https://anonymous.4open.science/r/Adversarial-Image-Captioning-Attack}.}.

\subsection{Datasets}

Our experiment was conducted on the COCO \cite{cite28} and Flickr8k \cite{cite34} datasets. Each image in the COCO dataset is accompanied by five human-generated captions, providing rich linguistic annotations that describe the visual content with varying levels of detail and perspectives. 
%Our experiment was conducted on the COCO dataset \cite{cite28}. This dataset for image captioning is a widely used benchmark in computer vision research. Each image in the dataset is accompanied by five human-generated captions, providing rich linguistic annotations that describe the visual content with varying levels of detail and perspective. Meanwhile, we have also selected the Flickr8k \cite{cite34} dataset. 
The Flickr8k dataset is sourced from the Flickr image-sharing platform. It includes 8,000 images, each accompanied by five distinct captions, resulting in 40,000 captions.
\subsection{Victim Models and Baselines}
We use two victim image captioning models with leading-edge performance to examine our attacking algorithm. They are ``Show, Attend, and Tell" (abbreviated as ``SAT" ) \cite{cite23} and ``BLIP" \cite{cite4}. For image captioning attack baselines, we chose ``Show and Fool" \cite{cite19} and ``GEM" \cite{cite33}. We also finetuned ``One Pixel Attack" \cite{cite17}, initially designed for fooling image classification models to generate adversarial images for comparison to our approach.
% \subsection{Evaluation Metrics}
\subsection{Metrics}
To examine and measure the performance of the attacks, we reported the attack performance of different methods using several metrics.
\subsubsection{BLEU Score}
{BLEU} (Bilingual Evaluation Understudy) score \cite{cite27} is a commonly used metric in the field of natural language processing, including image captioning. In our experiment, we used BLEU-1 and BLEU-2 (unigram- and bigram-based BLEU scores) and BLEU-4 to deal with four-word phrases for longer captions. The equation of BLEU-4 is shown in Equation~\ref{equation: BLEU}:
\begin{equation}
\label{equation: BLEU}
\begin{split}
\text{BLEU-4} &= \text{BP} \times \exp\left(\frac{1}{4} \sum_{n=1}^{4} \log \left(\text{precision}_n\right)\right), 
\end{split}
\end{equation}

\noindent where $\text{precision} = \frac{Number\ of\ correct\ word\ tokens\ generated}{Number\ of\ total\ word\  tokens\ generated}$. Sometimes, candidates might be very small for longer captions and missing important information relative to the reference. So we include the Brevity Penalty (BP) to penalize predicted captions that are too short compared to the reference captions. The BP is defined as follows:
\begin{equation} 
\begin{split}
\mathrm{BP} = \begin{cases}1 & \text {if } c>r \\ e^{(1-r / c)} & \text {if } c \leq r\end{cases}
\end{split}
\end{equation}
where $r$ refers to the length of the original caption, and $c$ refers to the length of the generated caption. 

\subsubsection{ROUGE Score}
\begin{figure*}[!t]
    \centering
    \includegraphics[width=0.7\textwidth]{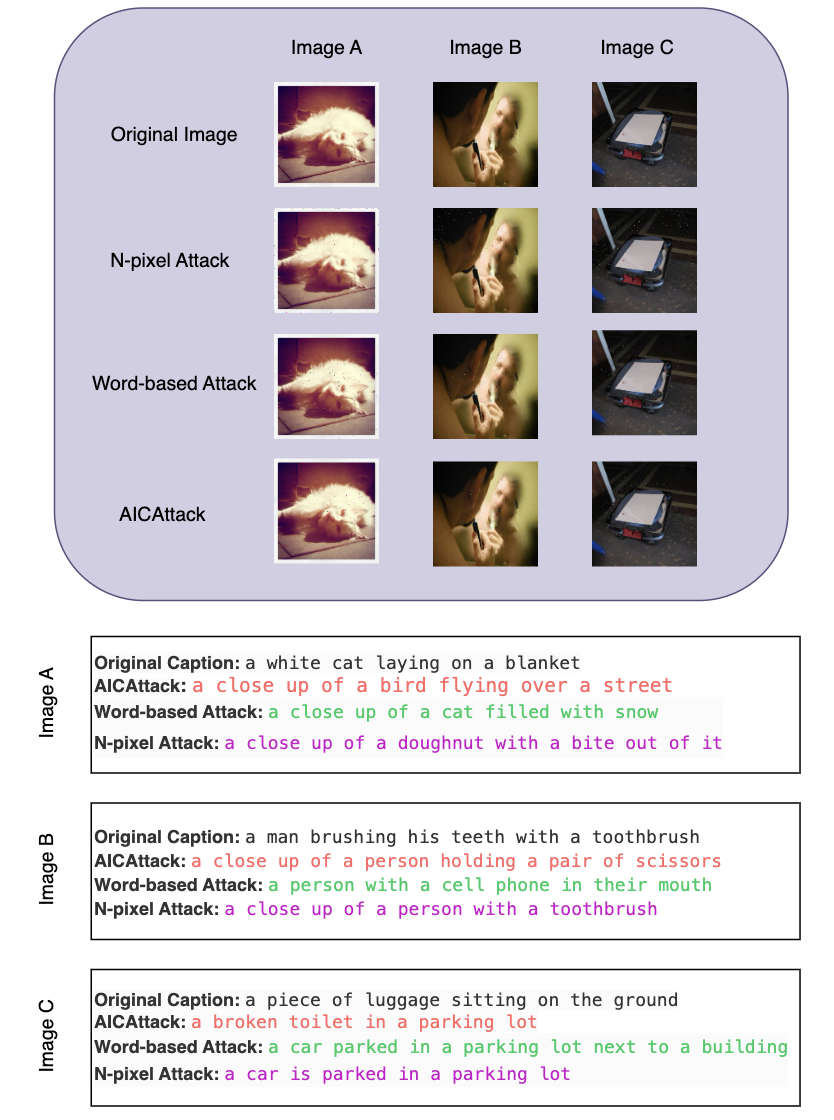}
    \caption{Visual examples illustrating different attack strategies, accompanied by captions.}
    \label{fig:examples}
\end{figure*}
\begin{table*}[!t]
    \centering
    \caption{The performance of two baseline victim models tested on COCO and Flickr8k datasets.}
    \begin{tabular}{cccccc}
        \toprule
        &&BLEU1&BLEU2&BLEU3&BLEU4\\
        \midrule
        \multirow{2}*{SAT}&COCO&71.8&50.4&36.7&25.0\\
        \cmidrule(lr){2-6}
         &Flickr8k&67.0&45.7&31.4&21.3\\
        \midrule
        \multirow{2}*{BLIP} &COCO&73.1&48.9&38.2&26.6\\
        \cmidrule(lr){2-6}
        &Flickr8k&70.1&47.2&32.5&22.8\\
        \bottomrule
    \end{tabular}
    \vspace{10pt}
    \label{tab: baseline results}
\end{table*}

By using BLEU only, it may not fully capture human language's semantic and contextual nuances, and it might not always correlate perfectly with human judgment. Hence, we also report ROUGE-n \cite{cite31}, which measures the number of matching n-grams between the model-generated captions and human-produced/ground-truth reference captions. Our experiments used unigram and bigram ROUGE scores (i.e., ROUGE-1 and ROUGE-2).
\begin{equation}
\label{Rouge-N}
%\begin{split}
    \text{ROUGE-n}=\frac{\sum_{\substack{S \in \text{ReferenceCaptions}}} \sum_{\substack{\text{n-gram} \in S}} \text{Count}_{\text{match}}(\text{n-gram})}{\sum_{\substack{S \in \text{ReferenceCaptions}}} \sum_{\substack{\text{n-gram} \in S}} \text{Count}(\text{n-gram})}
%\end{split}
\end{equation}

\subsubsection{BR-measure}
% To examine and measure the performance of our approach, we reported unigram and bigram BLEU \cite{cite27} and ROUGE \cite{cite31} scores, where we use BLEU1 and ROUGE1 to represent the unigram scores and BLEU2 and ROUGE2 to represent bigram scores. 
Besides reporting BLEU and ROUGE individually to comprehensively represent the results, we introduced a new measure to simplify the process of evaluating attack results, which integrates the ROUGE and BLEU scores in a way similar to driving F-measure from precision and recall, which we call BR-measure:
\begin{equation}
    % \begin{split}
        \text{BR-measure} = \frac{BLEU * ROUGE}{BLEU + ROUGE}\\
    % \end{split}
\end{equation}
The BR-measure has a desirable property where the value of the BR-measure is high if only both BLUE and ROUGE values are high. The BR-measure will be low if the BLUE or the ROUGE value is low.

Before examining our AICAttack, we report the original captioning performance of two victim models on the COCO and Flickr8k datasets in Table~\ref{tab: baseline results}. 

\subsection{Experiment Analysis}\label{experiment analysis}

Our experiments were conducted by the following settings (the evaluation of the tuning of these parameters is studied in the later part of this section): attention region $k$ at 50\% (attacking pixels whose attention weight is above the median of all weights), with $\pm 50$ range for $s$ (i.e., modify at most 50-pixel intensity values for each attack pixel), parameter $\lambda$ is set to be 0.5 and targeting 500 to 1000 random image-caption pairs from the test dataset. For the ``SAT" model, we attack 500 pixels. Considering the larger image size in the ``BLIP" model, we extended the attack to 1000 pixels.

We demonstrate some attack outcomes in Fig.~\ref{fig:examples}. The results in Tables~\ref{tab:overall_table_BLIP} and~\ref{tab:overall_table_SAT} highlight our attack strategies' effectiveness across models. Notably, our ``AICAttack" (i.e., the ``Sentence-based Attack") method outperformed baseline approaches. Particularly, our AICAttack outperformed GEM and Show and Fool, which revealed the effectiveness of our ``Sentence-based Attack" work. 
\begin{landscape}   
\begin{table*}[!t]
    \centering
     \caption{The table presents the outcomes of our attack methods applied to BLIP with 1000 randomly selected samples from the COCO and Flickr8k datasets. All measures in the table denote the differences before and after the attacks (i.e., the value dropped after the attacks). N-Pixel Attack randomly selected pixels without using attention. Optimal outcomes are denoted in bold.}
    %\resizebox{\textwidth}{!}{%
    \begin{tabular}{cccccccc}
        \toprule
        \multirow{2}{*}{Datasets}&\multirow{2}*{Methods}&\multicolumn{6}{c}{Value Dropped After the Attack (Higher is Better)}\\
        \cmidrule(lr){3-8}
        &&BLEU1&BLEU2&BLEU4&ROUGE1&ROUGE2&{\shortstack{BR}}\\
        \hline
        \multirow{6}{*}{COCO}&One Pixel Attack&0.005&0.003&0.002&0.002&0.004&0.001\\
        \cmidrule(lr){2-8}
        &Show and Fool &0.054&0.086&0.033&0.006&0.038&0.005\\
        \cmidrule(lr){2-8}
        &GEM &0.051&0.093&0.048&0.006&0.034&0.004\\
        \cmidrule(lr){2-8}
        &N-Pixel Attack& 0.059 & 0.073 & 0.062& 0.006 & 0.020 & 0.032\\
        \cmidrule(lr){2-8}
        &Word-based Attack & \textbf{0.066} & 0.074 & 0.053&0.005 & 0.021 & 0.032\\
        \cmidrule(lr){2-8}
        &AICAttack \textbf{(ours)} & 0.060 & \textbf{0.104} & \textbf{0.066}& \textbf{0.008} & \textbf{0.041} & \textbf{0.005}\\
        \hline
        \multirow{6}*{Flickr8k}&One Pixel Attack  &0.003&0.002&0.004&0.004&0.001&0.002\\
        \cmidrule(lr){2-8}
        &Show and Fool&0.048&0.08&0.038&0.007&0.025&0.031\\
        \cmidrule(lr){2-8}
        &GEM&0.053&0.079&0.041&0.007&0.028&0.031\\
        \cmidrule(lr){2-8}
        &N-Pixel Attack &0.053&0.073&0.044&0.006&0.024&0.023\\
        \cmidrule(lr){2-8}
        &Word-based Attack & \textbf{0.057} & 0.071 &0.048& 0.006 & 0.022 & 0.032\\
        \cmidrule(lr){2-8}
        &AICAttack \textbf{(ours)}& 0.057 & \textbf{0.081} &\textbf{0.052}& \textbf{0.007} & \textbf{0.028} & \textbf{0.033}\\
        \bottomrule
    \end{tabular}
    \label{tab:overall_table_BLIP}
\end{table*}
\end{landscape}

\begin{landscape}
\begin{table*}[!t]
    \centering
     \caption{The table presents the outcomes of our attack methods applied to SAT with 1000 randomly selected samples from the COCO and Flickr8k datasets. All measures in the table denote the differences before and after the attacks (i.e., the value dropped after the attacks). N-Pixel Attack randomly selected pixels without using attention. Optimal outcomes are denoted in bold.}
    %\resizebox{\textwidth}{!}{%
    \begin{tabular}{cccccccc}
        \toprule
        \multirow{2}*{Datasets}&\multirow{2}*{Methods}&\multicolumn{6}{c}{Value Dropped After the Attack (Higher is Better)}\\
        \cmidrule(lr){3-8}
        &&BLEU1&BLEU2&BLEU4&ROUGE1&ROUGE2&{\shortstack{BR}}\\
        \hline
        \multirow{6}*{COCO}&One Pixel Attack  &0.009&0.004&0.003&0.005&0.002&0.002\\
        \cmidrule(lr){2-8}
        &Show and Fool&0.119&0.163&0.109&0.037&0.045&0.063\\
        \cmidrule(lr){2-8}
        &GEM &0.120&0.177&0.083&0.039&0.049&0.061\\
        \cmidrule(lr){2-8}
        &N-Pixel Attack & 0.125 & 0.179 &0.130& 0.074 & 0.070 & 0.064\\
        \cmidrule(lr){2-8}
        &Word-based Attack & 0.117 & 0.170 &0.127& 0.070 & 0.067 & 0.062\\
        \cmidrule(lr){2-8}
        &AICAttack \textbf{(ours)} & \textbf{0.127} & \textbf{0.188} &\textbf{0.131}& \textbf{0.075} & \textbf{0.074} & \textbf{0.065}\\
        \hline
        \multirow{6}*{Flickr8k}&One Pixel Attack &0.004&0.003&0.002&0.004&0.003&0.004\\
        \cmidrule(lr){2-8}
        &Show and Fool&0.104&0.153&0.014&0.030&0.037&0.043\\
        \cmidrule(lr){2-8}
        &GEM&0.102&0.151&0.013&0.033&0.048&0.045\\
        \cmidrule(lr){2-8}
        &N-Pixel Attack&0.101&0.166&0.010& 0.032&0.061&0.043\\
        \cmidrule(lr){2-8}
        &Word-based Attack & 0.103 & 0.168 &0.107& 0.048 & 0.066 & 0.045\\
        \cmidrule(lr){2-8}
        &AICAttack \textbf{(ours)} & \textbf{0.114} & \textbf{0.175} &\textbf{0.108}& \textbf{0.053} & \textbf{0.068} & \textbf{0.049}\\
        \bottomrule
    \end{tabular}
    \label{tab:overall_table_SAT}
\end{table*}
\end{landscape}
The comparison between ``Sentence-based Attack" and ``Word-based Attack" methods exposed a more pronounced decidable in the former. This distinction arises from the ``Word-based Attack" approach's attention selection, contrasting the more focused nature of ``Sentence-based" selection. The sentence exemplifies this distinction ``a cat is sitting in a bathroom sink," wherein ``Word-based Attack" attends to ``a", ``in" and ``is". Consequently, a broader region from the image, encompassing non-significant elements, was incorporated into the candidate region. As visualization in Fig.~\ref{fig:attention in step}, Word-based selection indicates a broader scope, incorporating boundary pixels. Conversely, the attention region depicted in Fig.~\ref{fig:attention_area} for Sentence-based selection is more confined, centering exclusively on pertinent entities.

\subsection{Ablation and Hyperparameters Studies}\label{Ablation Studies}
\begin{figure}[!t]
  \centering
      \includegraphics[width=0.9\textwidth]{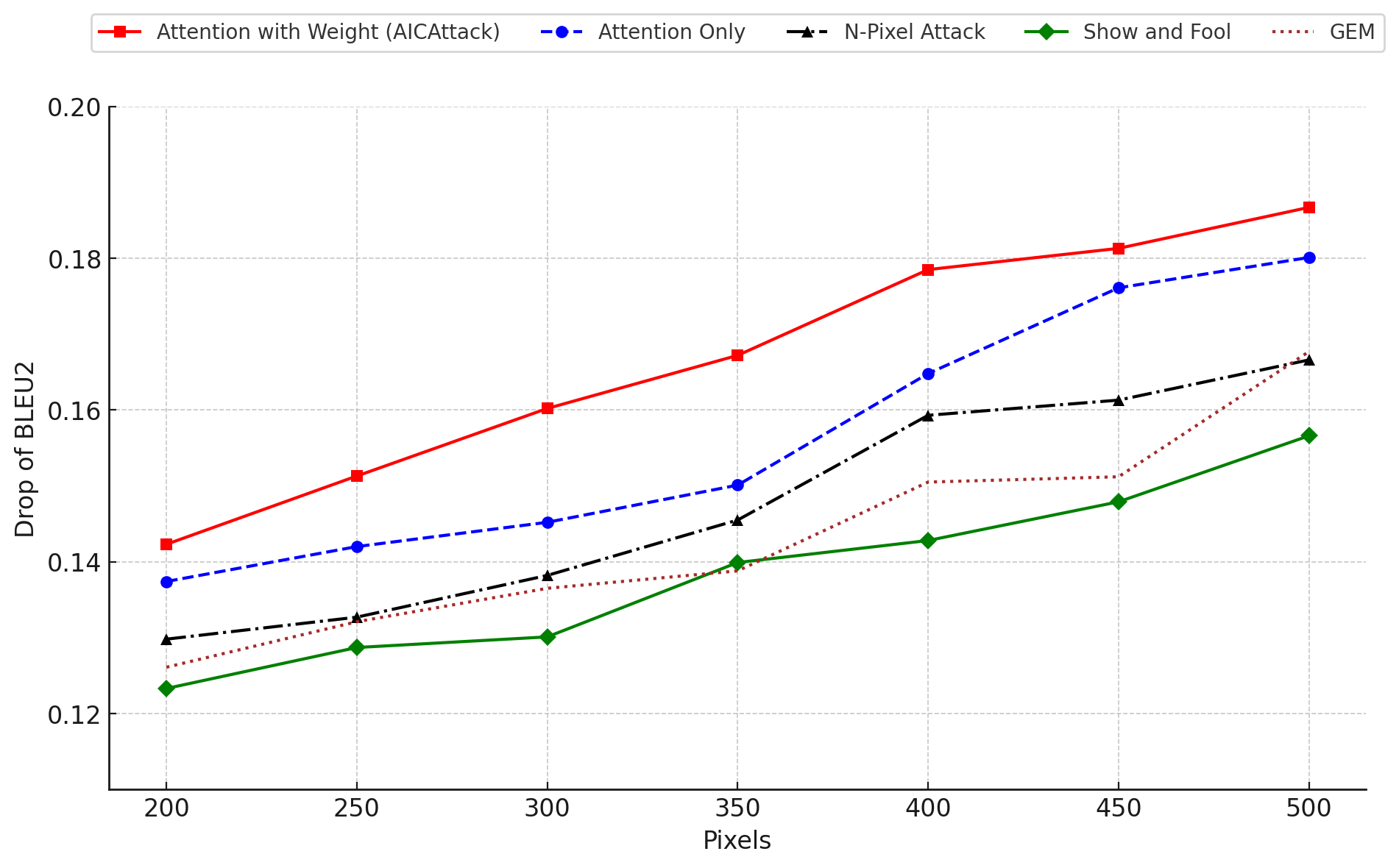}
  \caption{Drops of BLEU2 scores before and after five attack scenarios across different pixel counts.}
  \label{pixels comparison}
\end{figure}

We introduce ablation experiments to validate the effectiveness of our AICAttack method, apart from employing baselines to substantiate attention. As shown in Fig.~\ref{pixels comparison}, the figure displays the variation in BLEU2 scores under five attack methods across different pixel counts. Firstly, compared to other baselines, it can be observed that our approach consistently maintains the optimal performance under extreme conditions (attack fewer pixels), our method consistently maintains the optimal performance. This is attributed to attention and weight selection capabilities that facilitate the choice of the most optimal pixels for attack. Furthermore, we can observe that the attention method combined with weight outperforms the attention-only approach.
\subsubsection{Analysis of Differential Evolution in AICAttack}
In this section, we study the robustness of Differential Evolution in our methods with three baseline optimization algorithms. The three baseline algorithms are: two gradient-based approaches - Projected Gradient Descent (PGD)~\cite{cite11} and Fast Gradient Sign Method (FGSM)~\cite{cite9} - along with Particle Swarm Optimization (PSO)~\cite{cite45}.
For both PGD (Projected Gradient Descent) and FGSM (Fast Gradient Sign Method), we generate adversarial images $\mathbf{x'}$ by perturbing the original image $\mathbf{x}$ that maximizes a specified loss function $\mathcal{L}$ (BLEU score decrease), while constraining the perturbations to attention region with a binary mask $M_{i}$, as shown in Equation~\ref{optimization}.

\begin{equation}
\begin{aligned}
M_{i} &= 
\begin{cases} 
1 & \text{if } \mathbf{x_i} \ \text{in} \ \text{attention region} \\
0 & \text{else}
\end{cases} \\
\mathbf{x'} &= \mathbf{x} + \epsilon \cdot M_{i} \odot \operatorname{sign}\left(\nabla_x \mathcal{L}\right)\\
\end{aligned}
\label{optimization}
\end{equation}
where:
\begin{itemize}
    \item $\operatorname{sign}\left(\nabla_x \mathcal{L}\right)$: The sign of the gradient.
    \item $\epsilon$: The perturbation magnitude is set as 8 in the experiment.
    \item $\odot$: The element-wise multiplication operator.
\end{itemize}
We adapt and fine-tune the Particle Swarm Optimization (PSO) algorithm to optimize perturbations by iteratively refining particle populations constrained within attention regions. Each particle represents a candidate perturbation vector that captures spatial and color dimensions. The PSO experiment is configured with the following parameters:
\begin{itemize}
    \item $\ell_{\infty}$-norm: $\epsilon_{\infty}=16$
    \item $\ell_{2}$-norm: $\epsilon_{2}=5.0$
    % \item $\epsilon=8$
    \item Swarm size (Number of particles) = 50
\end{itemize}
Our experimental results demonstrate that PGD and FGSM require 30 iterations to maximize the BLEU score difference between reference and adversarial captions for effective attacks. PSO requires 50 iterations to achieve comparable performance. As shown in Table~\ref{tab:optimization_results}, our DE algorithm in AICAttack achieves superior performance with only 5 iterations, significantly outperforming other optimization methods and demonstrating its effectiveness in adversarial optimization.
\begin{table}[!t]
\centering
\caption{Comparative performance of optimization methods in adversarial attacks on SAT and BLIP models using BLEU and ROUGE metrics on COCO dataset.}
\label{tab:optimization_results}
\begin{tabular}{clcccccc}
\toprule
Model & Method & BLEU1 & BLEU2 & BLEU4 & ROUGE1 & ROUGE2 \\
\midrule
\multirow{4}{*}{SAT} 
    & FGSM      & 0.073   & 0.103   & 0.078   & 0.033   & 0.038   \\
    & PGD       & 0.081   & 0.122   & 0.093   & 0.054   & 0.062   \\
    & PSO       & 0.097   & 0.135   & 0.113   & 0.067   & 0.045   \\
    & DE (ours) & \textbf{0.127}   & \textbf{0.188}   & \textbf{0.131}   & \textbf{0.075}   & \textbf{0.074}   \\
\midrule
\multirow{4}{*}{BLIP} 
    & FGSM      & 0.021   & 0.056   & 0.031   & 0.006   & 0.007   \\
    & PGD       & 0.019   & 0.042   & 0.042   & 0.007   & 0.039   \\
    & PSO       & 0.036   & 0.077   & 0.051   & 0.007   & 0.040   \\
    & DE (ours) & \textbf{0.060}   & \textbf{0.104}   & \textbf{0.066}   & \textbf{0.008}   & \textbf{0.041}   \\
\bottomrule
\end{tabular}
\end{table}

\subsubsection{Number of Iterations in Genetics Optimization versus Attacking Performance}
\begin{figure}[!t]
  \centering
      \includegraphics[width=0.9\textwidth]{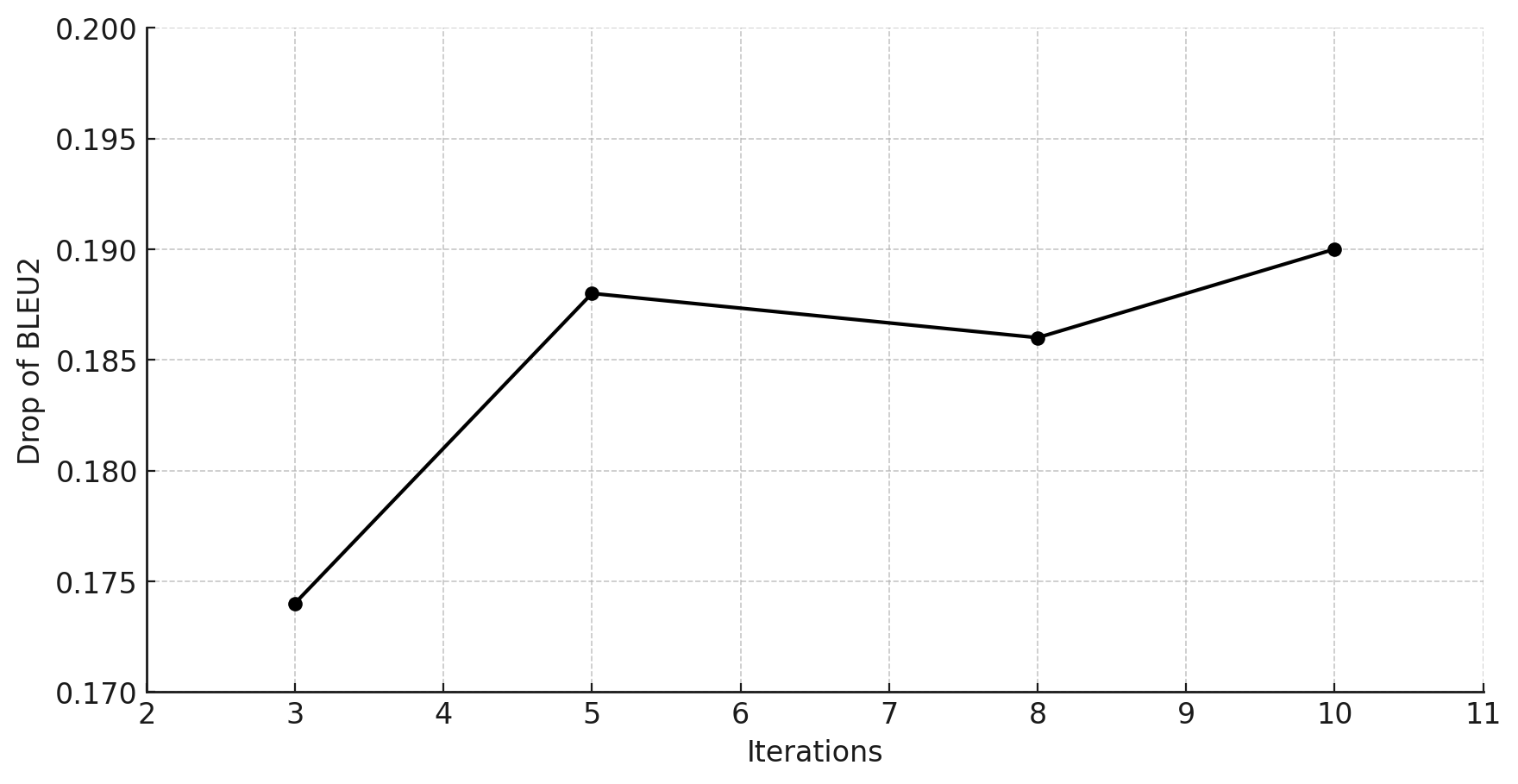}
      \caption{Drops in BLEU2 scores across varying iteration counts in the differential evolution algorithm.}

  \label{number of iterations}
\end{figure}

To examine the impact of iteration numbers on performance, we subject 1000 samples from the COCO test set to our AICAttack method under various iteration configurations. The results are shown in Fig.~\ref{number of iterations}. The attack performance notably improves when increasing from three to five iterations. However, beyond five iterations, up to ten, the BLEU 2 score fluctuates, indicating that our methodology achieves stable performance with iterations exceeding five.

\subsubsection{Candidate Region Analysis}
\begin{figure}[!t]
  \centering
  \includegraphics[width=0.9\textwidth]{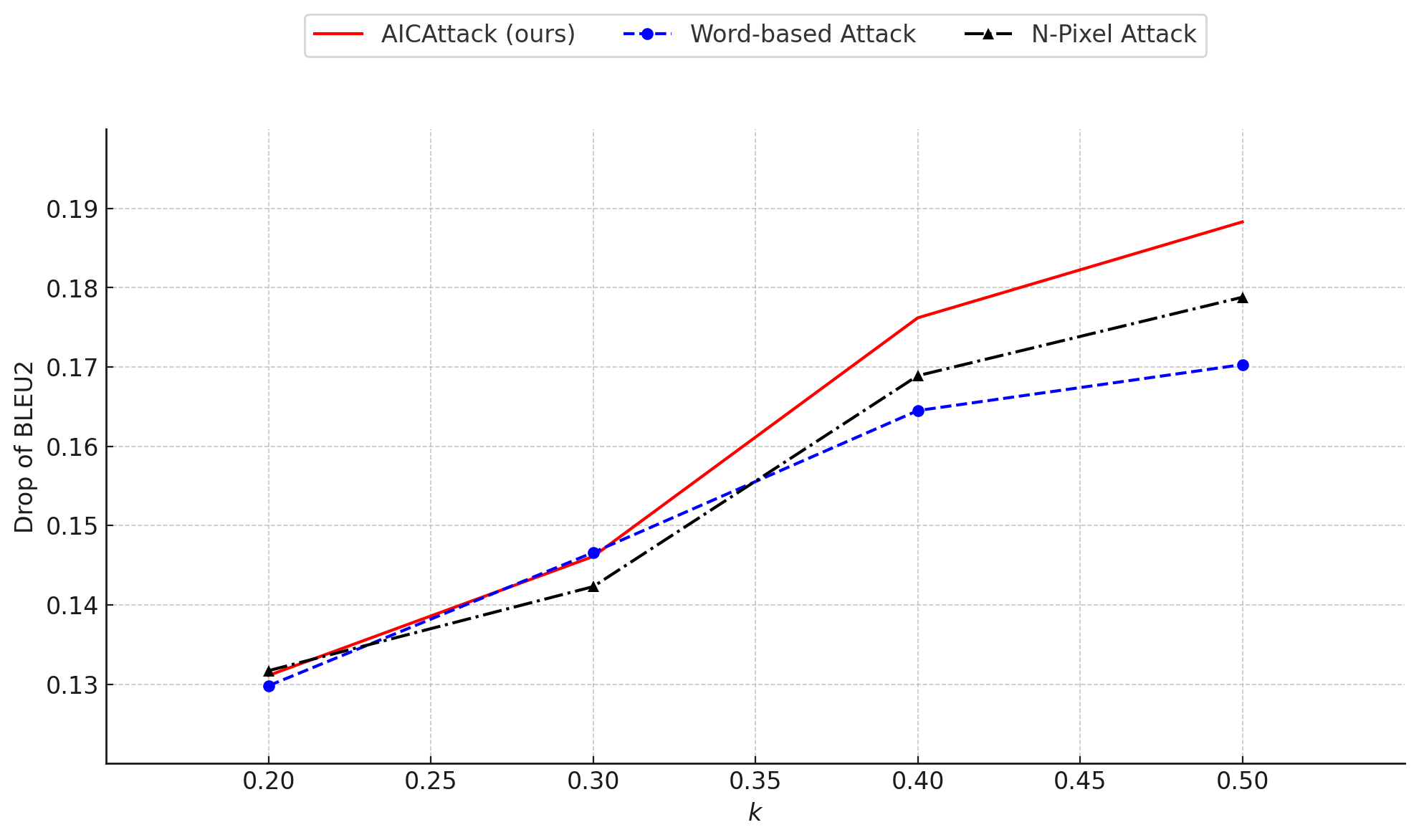}
  \vspace{0.5cm}
  \caption{Drops of BLEU2 scores before and after attack when applying multiple attention regions $k$.}
  \label{k comparison}
\end{figure}

We examined the impact of varying candidate region $k$ sizes on experimental outcomes, shown in Fig.~\ref{k comparison}. Note that this consideration differs from pixel count. 

% \subsubsection{choice of hyperparameter $\lambda$ in genetics optimization }

% \subsubsection{Comparison with different evaluation metrics}

\begin{figure}[!t]
  \centering
  \includegraphics[width=0.9\textwidth]{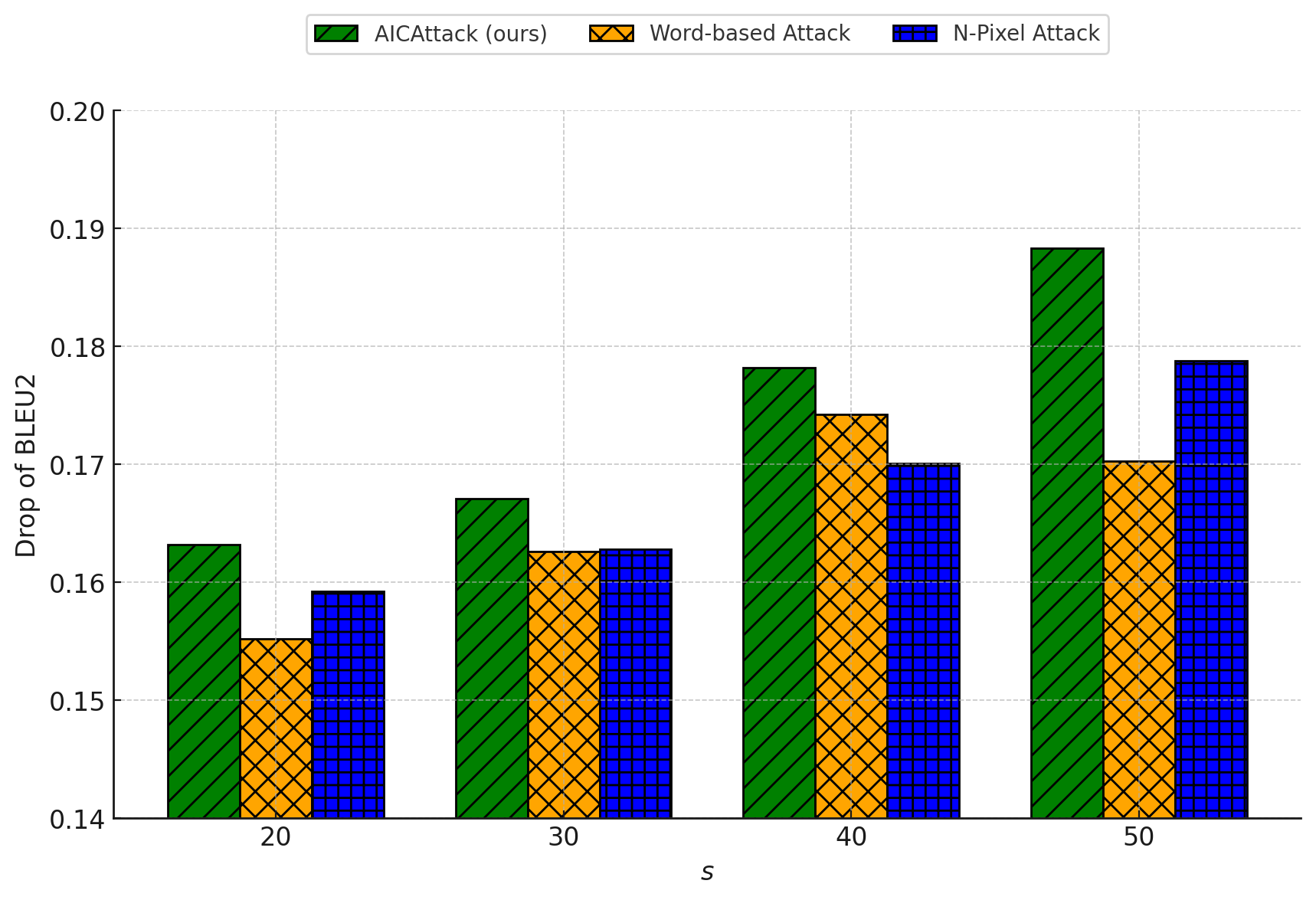}
  \caption{Drops of BLEU2 scores before and after attack when applying multiple attack strengths $s$.}
  \label{attack strength}
\end{figure}

\begin{figure}[!t]
  \centering
   \includegraphics[width=0.9\textwidth]{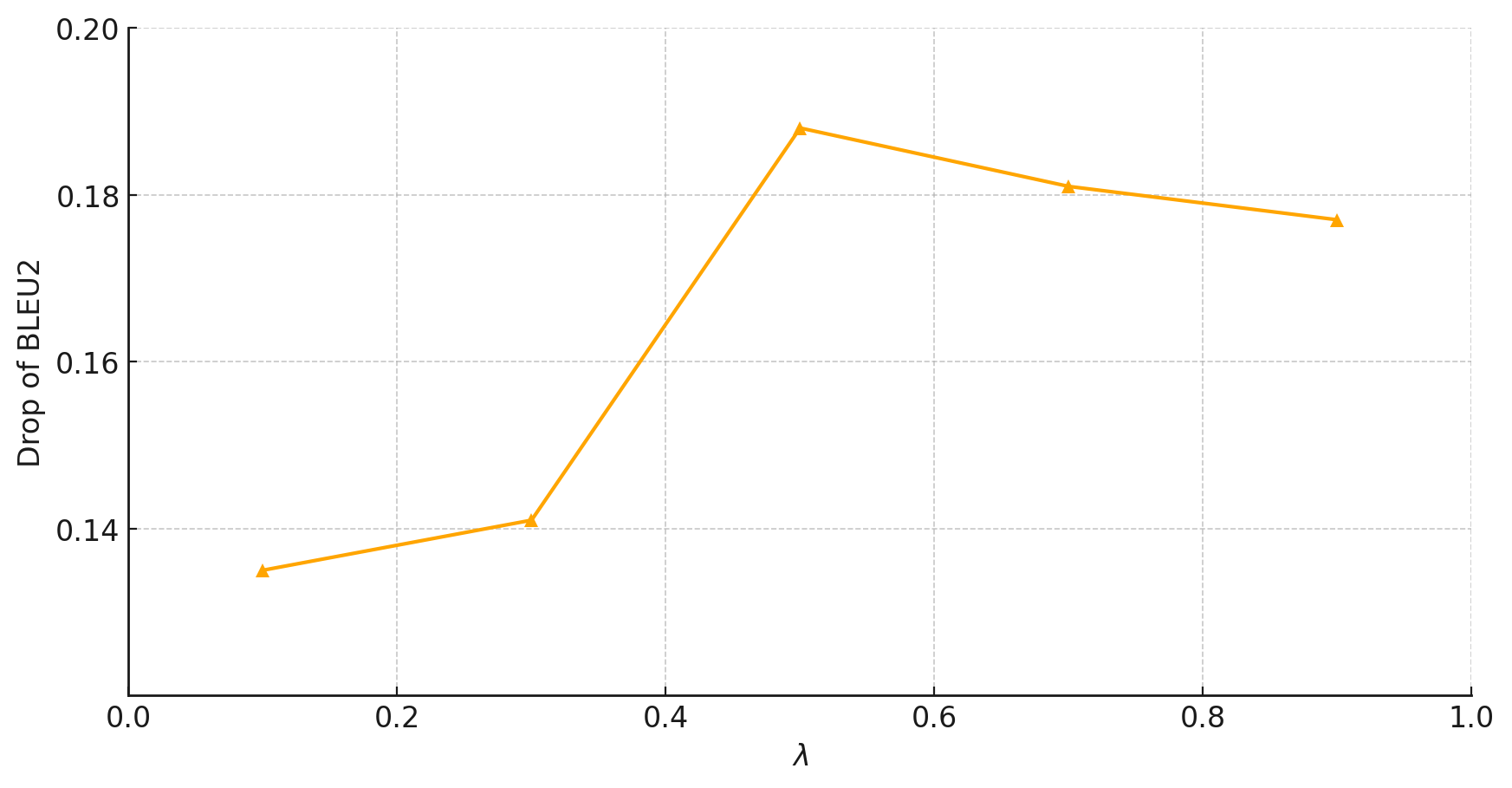}
  \caption{Drops in BLEU2 scores across varying $\lambda$ counts in the differential evolution algorithm.}
  \label{variation of lambda}
\end{figure}

The value of $k$ determines the initial size of the candidate region, while the number of pixels dictates how many are selected for attacks within this area. As shown in the graph, fluctuations in both directions occur as $k$ varies, but the overall trend is predominantly upward. This suggests that when the attention region is too narrow, the pixels targeted for attack may miss critical information. The candidate region must expand to a critical threshold before effectively capturing relevant sensitive data. This pattern underscores the importance of selecting an appropriate range for $k$ to ensure comprehensive coverage of vulnerable areas in the input.
\subsubsection{Attack Strength Analysis}\label{Attack Strength Analysis}
On the other hand, we explore the impact of different attack intensities (strength $s$) on the results. It can be observed in Fig.~\ref{attack strength} that in our proposed method, larger intensities lead to more noticeable changes in BLEU scores. This can be attributed to a significant alteration in pixel colouration, which impacts the model's capacity to interpret the contents of the image accurately.
\subsubsection{Scale Factor Analysis}\label{Scale Factor Analysis}
Finally, we conduct scale factor $\lambda$ experiments in the differential evolution algorithm. The result is shown in Fig.~\ref{variation of lambda}. A larger value of $\lambda$ enables the algorithm to conduct a wider search across the solution space. This can help avoid local minima, although it carries the risk of instability or not achieving accurate solutions. Conversely, a smaller $\lambda$ value fosters exploitation, concentrating the search in a more confined area. This can be advantageous for detailed adjustments but may lead to the algorithm becoming trapped in local optima. Hence, in our AICAttack, we pick 0.5 between 0 to 1 as the $\lambda$ value.

\begin{figure}[t]
\centering
   \includegraphics[width=0.9\textwidth]{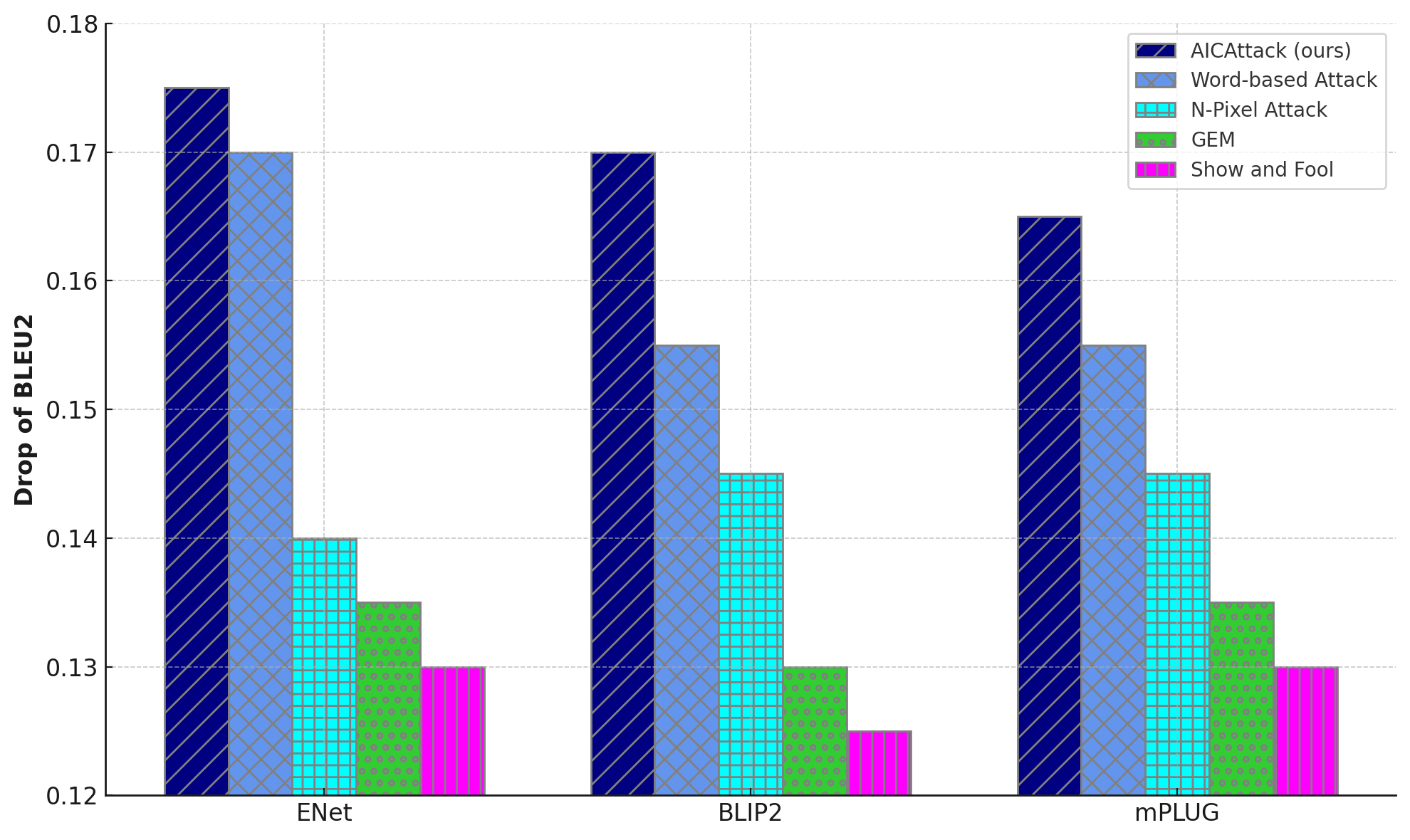}
\caption{Drops of BLEU2 scores reported on multiple baseline captioning models with COCO datasets.}
\label{transfer attack COCO}
\end{figure}

\begin{figure}[!t]
       \includegraphics[width=0.9\textwidth]{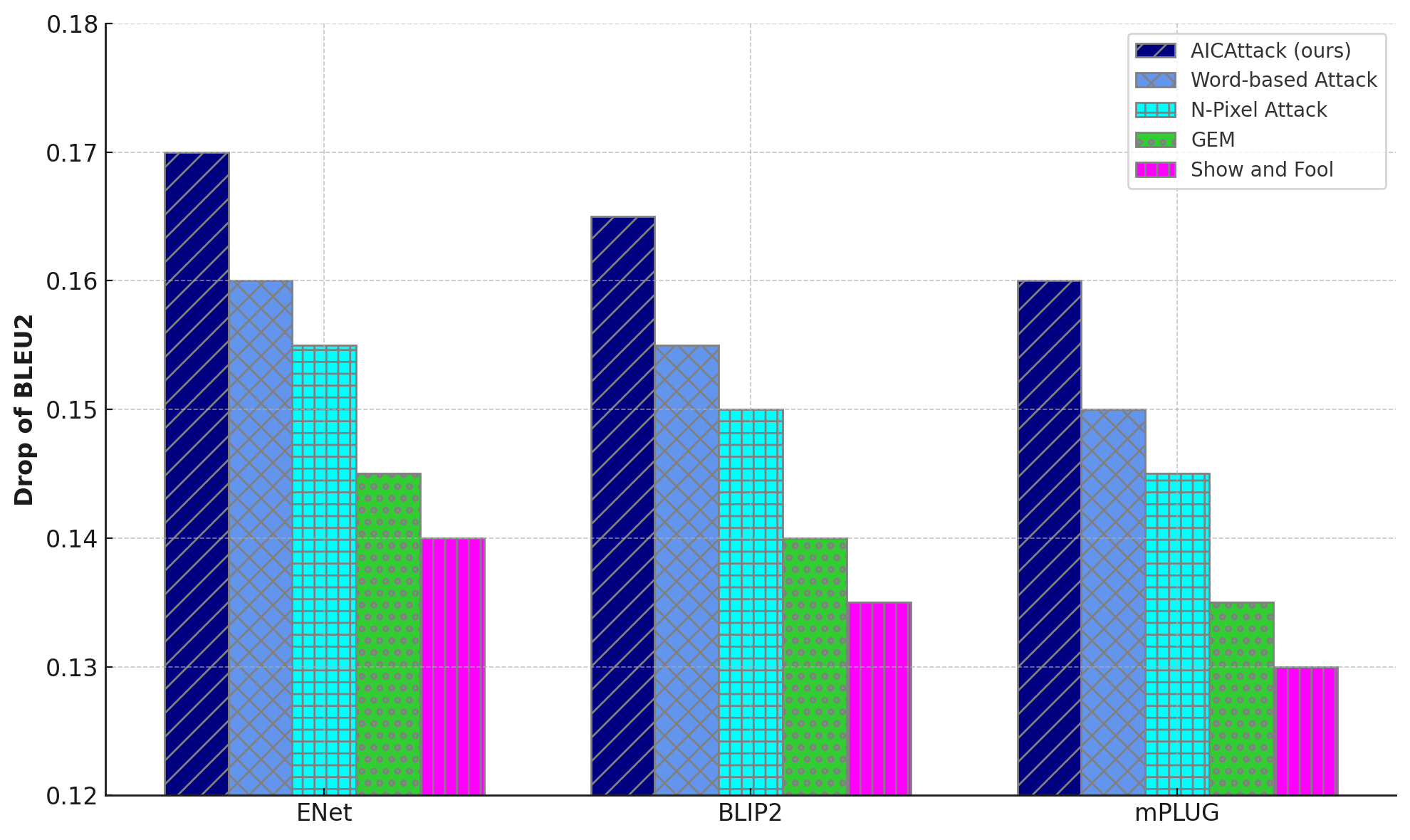}
\caption{Drops of BLEU2 scores reported on multiple baseline captioning models with Flickr8k datasets.}
\label{transfer attack filkr8k}
\end{figure}

% This is attributed to selecting non-essential pixels in random attacks, which does not influence the outcomes. These further underscores the effectiveness of our attention mechanism.
\subsection{Transferability of Attacks} \label{Transferability of Attacks}
To test our model's reaction to unknown captioning models $F'$, we conducted a set of experiments with three baseline captioning works mPLUG\cite{cite35}, BLIP2\cite{cite36}, ExpansionNet v2 \cite{cite37}(denote as ENet). Specifically, we selected adversarial examples designed for SAT to attack baselines across different attack methods. As results shown in Fig.~\ref{transfer attack COCO} and Fig.~\ref{transfer attack filkr8k}, our AICAttack generates adversarial examples with higher transferability among five attacking approaches.

% \label{Transfer Attack}
%~\ref{named}

\subsection{Adversarial Retraining}\label{Adversarial Retraining}
\begin{table*}[!t]
    \centering
    \caption{Adversarial retraining results on different scenarios. All the increases and decreases are based on training results with $100\%$ or $5\%$ training data, respectively.}
    \begin{tabular}{lcccc}
    \toprule
        Data for training&BLEU1&BLEU2&BLEU4&Time(h/epoch)\\
        \midrule
         $100\%$ training data &71.824&50.381&25.033&1.42\\
         \ \ \ + 1000 adversarial&-0.013&-0.017&-0.06&1.54\\
          \ \ \ + 2000 adversarial&-0.003&-0.08&+0.05&1.67\\
          \ \ \ + 3000 adversarial&+0.027&+0.033&+0.14&1.83\\
          \ \ \ + 5000 adversarial&+0.063&+0.052&+0.042&2.04\\
         $5\%$ training data &56.333&51.387&23.667&0.06\\
         \ \ \ + 100 adversarial&+3.251&+1.33&-0.216&0.06\\
         \bottomrule
    \end{tabular}
    \vspace{10pt}
    \label{tab: adversarial retraining}
\end{table*}

\begin{table*}[!t]
    \centering
    \caption{Attack performance on pre-trained SAT in different scenarios as shown in Table~\ref{tab: adversarial retraining}.}
    \begin{tabular}{lccc}
    \toprule
        Retraining Settings&BLEU1&BLEU2&BLEU4\\
        \midrule
         100\% training data& -59.124&-31.581&-11.533\\
         \ \ \ 1000 adversarial&-58.273&-30.333&-9.175\\
         \ \ \ 2000 adversarial&-58.031&-29.867&-9.011\\
         \ \ \ 3000 adversarial&-57.336&-29.732&-8.767\\
         \ \ \ 5000 adversarial&-56.731&-28.667&-7.913\\
         5\% training data &-21.592&-19.033&-13.833\\
         \ \ \ 100 adversarial&-4.261&-5.012&-1.417\\
    \bottomrule
    \end{tabular}
    \label{tab:retraining attack}
\end{table*}
This section examines AICAttack's potential to enhance victim model's BLEU scores. We initially trained the Show, Attend and Tell (SAT) model on both the full COCO training set and a 5\% subset. Subsequently, we generated sets of 100, 1000, 2000, 3000, and 5000 adversarial examples using AICAttack. For adversarial retraining, we created new training sets by combining the original data with these adversarial examples. Specifically, we added 1000, 2000, 3000, and 5000 adversaries separately to the full training data and 100 adversaries to the 5\% training data. This approach allows us to evaluate the impact of adversarial retraining across different scales of both original and adversarial data.
The training followed the primary settings described in the original paper \cite{cite23} and its implementation\footnote{The implementation of Show, Attend and Tell can be found here: \url{https://github.com/kelvinxu/arctic-captions}.}. To accelerate training, we employed a pre-trained ResNet-101 \cite{cite3} from PyTorch's torchvision module and focused on training only the decoder. We used an initial learning rate of $4\times10^{-4}$ with the Adam optimizer. An early stopping mechanism was implemented, terminating training when the BLEU-4 score began to decline, even if the loss continued to decrease. Our experiment was conducted on a system with two NVIDIA L40 GPUs, each featuring 48 GB memory.

\subsubsection{The Accuracy of Retrained Model}\label{The Accuracy of Retrained Model}
As shown in Table~\ref{tab: adversarial retraining}, when training with all training data and 1000 adversarial samples, the BLEU scores decrease by 0.013, 0.017 and 0.06, respectively. However, when more adversaries are added (2000, 3000, 5000), the BLEU scores improve consistently. Meanwhile, the BLEU scores improve under low-data scenarios (5\% training data with 100 adversaries added). This result shows that adding an adequate amount of new adversarial examples to the original training data leads to better outcomes. Regarding training time consumption, we computed training time excluding time for generating adversarial samples. The retraining time increases as more adversarial examples are added, especially for 100\% training data conditions. However, for the small subset, adding only 100 adversarial examples does not increase the training time. For large datasets, although the time for retraining is increased, that cost is well worth because it ensures the model can be more robust against adversarial attacks.

\subsubsection{The Robustness Confronting Adversarial Attacks}
To measure the effectiveness of the model's robustness after adversarial retraining, we use our AICAttack to foul each retrained model checkpoint in Table~\ref{tab: adversarial retraining}. In Table~\ref{tab:retraining attack}, the change in the BLEU score shows that adversarial training makes the attack less effective, with fewer BLEU scores dropping, and the more adversarial examples are added during retraining, the more robust the model becomes to attacks. This phenomenon is more significant in the low-resource scenario due to training with little data. These results suggest that AICAttack can be used to improve retrained captioning models’ robustness with a considerable BLEU score drop.

\section{Conclusion and Future Work}
In this research, we introduced AICAttack, a robust and versatile adversarial learning strategy for the attack of image captioning models. 
Our black-box approach harnesses the power of an attention mechanism and differential evolution optimization to orchestrate subtle yet effective pixel perturbations. It avoids the complex extraction of parameters from encoder-decoder models while keeping the attack cost within a minimal range. Another critical innovation of AICAttack is its attention-based candidate selection mechanism, which identifies optimal pixels for perturbation, enhancing the precision of our attacks.
Through extensive experimentation on benchmark datasets and captioning models, we have demonstrated the superiority of AICAttack in achieving significantly higher attack success rates compared to state-of-the-art methods. For the transferability of our AICAttack, we have fine-tuned it for Visual Question Answering (VQA) tasks in our future works, and we also plan to develop defensive strategies against image captioning attacks by designing more robust learning algorithms for image captioning models.

\section{Statements and Declarations}
The authors declared that they have no conflicts of interest in this work.

% \section{Declarations}
% \subsection{Availability of Data and Materials}
% The COCO dataset is available from \url{https://cocodataset.org/#home}.
% The Flickr8k dataset is available from \url{https://www.kaggle.com/datasets/adityajn105/flickr8k}. 
% Two victim image captioning models ``Show, Attend, and Tell'' and ``BLIP'' are available from \url{https://github.com/AaronCCWong/Show-Attend-and-Tell} and \url{https://github.com/salesforce/BLIP}. 
% Image captioning attack baselines ``Show and Fool'', ``GEM'' and ``One Pixel Attack'' are available from \url{https://github.com/huanzhang12/ImageCaptioningAttack}, \url{https://github.com/wubaoyuan/adversarial-attack-to-caption/tree/master} and \url{https://github.com/Hyperparticle/one-pixel-attack-keras}.

\bibliographystyle{unsrt}

\bibliography{bibiloiography.bib}
\end{document}